%% file: cvla.tex
\documentclass[sigconf]{acmart}

\usepackage{bm}
\usepackage{array}
\usepackage{float}
\usepackage{pifont}
\usepackage{lipsum}
\usepackage{xcolor}
\usepackage{balance}
\usepackage{colortbl}
\usepackage{latexsym}
\usepackage{graphicx}
\usepackage{makecell}
\usepackage{multirow}
\usepackage{enumitem}
\usepackage{booktabs}
\usepackage{bibentry}
\usepackage{hyperref}
\usepackage{subfigure}
\usepackage{nicematrix}
\usepackage[english]{babel}
\usepackage[export]{adjustbox}
\newcommand{\ms}[2]{{#1\tiny{$\pm$#2}}}
\newcommand{\bms}[2]{{\textbf{#1}\tiny{$\pm$#2}}}
\definecolor{MyRed}{HTML}{DC3977}
\definecolor{MyGreen}{HTML}{089099}

\AtBeginDocument{%
  }

\copyrightyear{2024}
\acmYear{2024}
\setcopyright{acmlicensed}
\acmConference[ICMR '24]{Proceedings of the 2024
International Conference on Multimedia Retrieval}{June 10--14, 2024}{Phuket,
Thailand}
\acmBooktitle{Proceedings of the 2024 International Conference on Multimedia
Retrieval (ICMR '24), June 10--14, 2024, Phuket, Thailand}
\acmDOI{10.1145/3652583.3658094}
\acmISBN{979-8-4007-0619-6/24/06}

\begin{document}

\title{Comment-aided Video-Language Alignment via Contrastive Pre-training for Short-form Video Humor Detection}

\author{Yang Liu}
\affiliation{
  \institution{Soochow University}
  \city{Suzhou}
  \country{China}
}
\email{yliucs@stu.suda.edu.cn}

\author{Tongfei Shen}
\affiliation{
  \institution{Soochow University}
  \city{Suzhou}
  \country{China}
}
\email{tfshen@stu.suda.edu.cn}

\author{Dong Zhang}
\authornote{Corresponding author.}
\affiliation{
  \institution{Soochow University}
  \city{Suzhou}
  \country{China}
}
\email{dzhang@suda.edu.cn}

\author{Qingying Sun}
\affiliation{
  \institution{Huaiyin Normal University}
  \city{Huai'an}
  \country{China}
}
\email{qingying.sun@foxmail.com}

\author{Shoushan Li}
\affiliation{
  \institution{Soochow University}
  \city{Suzhou}
  \country{China}
}
\email{lishoushan@suda.edu.cn}

\author{Guodong Zhou}
\affiliation{
  \institution{Soochow University}
  \city{Suzhou}
  \country{China}
}
\email{gdzhou@suda.edu.cn}

\renewcommand{\shortauthors}{Yang Liu et al.}

\begin{abstract}
\input{section/1-abs}
\end{abstract}

\begin{CCSXML}
<ccs2012>
   <concept>
       <concept_id>10002951</concept_id>
       <concept_desc>Information systems</concept_desc>
       <concept_significance>500</concept_significance>
       </concept>
   <concept>
       <concept_id>10002951.10003317.10003347.10003353</concept_id>
       <concept_desc>Information systems~Sentiment analysis</concept_desc>
       <concept_significance>500</concept_significance>
       </concept>
   <concept>
       <concept_id>10010147.10010178.10010179</concept_id>
       <concept_desc>Computing methodologies~Natural language processing</concept_desc>
       <concept_significance>300</concept_significance>
       </concept>
 </ccs2012>
\end{CCSXML}

\ccsdesc[500]{Information systems}
\ccsdesc[500]{Information systems~Sentiment analysis}
\ccsdesc[300]{Computing methodologies~Natural language processing}

\keywords{Humor Detection, Short-form Video, Dataset, Interactive Comments, Video-Language Alignment, Contrastive Pre-Training}


\maketitle

\input{section/2-intro}
\input{section/3-related_work}
\input{section/4-method}
\input{section/5-experiments}
\input{section/6-conclusion}

\begin{acks}
This work was supported by NSFC grants (No. 62206193 and No. 62006093), and Opening Foundation of Jiangsu Big Data Intelligent Engineering Laboratory of Soochow University (Grant No. SDGC2158).
\end{acks}

\bibliographystyle{ACM-Reference-Format}
\balance
\bibliography{cvla}

% \appendix









\end{document}

%% file: section/1-abs.tex
The growing importance of multi-modal humor detection within affective computing correlates with the expanding influence of short-form video sharing on social media platforms.
In this paper, we propose a novel two-branch hierarchical model for short-form video humor detection (SVHD), named \textbf{C}omment-aided \textbf{V}ideo-\textbf{L}anguage \textbf{A}lignment (\textbf{CVLA}) via data-augmented multi-modal contrastive pre-training.
Notably, our CVLA not only operates on raw signals across various modal channels but also yields an appropriate multi-modal representation by aligning the video and language components within a consistent semantic space.
The experimental results on two humor detection datasets, including DY11k and UR-FUNNY, demonstrate that CVLA dramatically outperforms state-of-the-art and several competitive baseline approaches.
Our dataset and code release at \href{https://github.com/yliu-cs/CVLA}{https://github.com/yliu-cs/CVLA}.

%% file: section/2-intro.tex
\section{Introduction}

Short-form video platforms such as DouYin (globally known as TikTok), Kwai, Snapchat, and Snack have risen as novel entertainment avenues, offering users the ability to view, create, comment on, and share captivating content typically lasting from a few seconds to several minutes.
The meteoric rise in these platforms' popularity has normalized the consumption of short-form videos as a quick respite during leisure time\cite{journals/chb/Wang20a}.
Accurately detecting the humorous aspects within these videos can greatly augment a platform's capacity to deliver targeted and tailored content recommendations that align with user preferences.
To this end, employing multi-modal approaches for the automatic detection of humor in short-form videos is imperative\cite{conf/emnlp/HasanRZZTMH19, conf/wacv/KayataniYOGCNT21}.

\input{figure/teaser}

However, we argue that short-form video humor detection (SVHD) currently faces at least three primary challenges:
1) The scale of unlabeled data currently utilized for pre-training is inadequate for ensuring the robustness of pre-trained models;
2) Conventional video humor detection approaches typically rely on pre-processed data features\cite{conf/aaai/HasanLR0MMH21, conf/ecai/LiuPZSLZ23}, where the additional step of feature extraction can introduce instability in data interpretation and increase time and resource expenditure;
3) The discrepancies may arise between the semantics of vision and language modalities.
For instance, as depicted in Figure~\ref{fig:teaser}, the video and title alone provide insufficient cues for humor prediction.
Moreover, the distinctive phrase \textit{pants all gone} significantly contributes to humor recognition in this short-form video.
Additionally, the presence of elements typically associated with non-humor, such as security guards and high-rise buildings, contrasts with the clear humorous cue provided by the phrase \textit{pants all gone} in the textual content.
This indicates that when the semantics of different modalities conflict, comments can often provide supplemental background knowledge that aids in the accurate interpretation of the video.
Consequently, we posit that aligning the semantics of multiple modalities as consistently as possible is crucial to integrate them from diverse modal perspectives into an appropriate multi-modal representation, thereby enhancing the accuracy of humor detection.

To address these issues, we first enlarge the unlabeled data scale for pre-training in DY24h\cite{conf/ecai/LiuPZSLZ23} to alleviate the dependency on annotation in SVHD task.
Then, we propose a novel hierarchical \textbf{C}omment-aided \textbf{V}ideo-\textbf{L}anguage \textbf{A}lignment (\textbf{CVLA}) approach by data-augmented contrastive learning for SVHD.
Note that our CVLA not only operates on raw signals across various modal channels but also yields an appropriate multi-modal representation by aligning the video and language components within a consistent semantic space.
Specifically, we first recast vision and audio as video branch information, and title and comments as language branch information.
Then, we tailor three Transformer encoders: two are used to encode the video branch and the language branch, separately.
The other is used to fuse the information of the two branches and yield the multi-modal representation for SVHD.
To align the semantics of the three representation, we propose a data-augmented contrastive pre-training strategy with large-scale unlabeled short-form videos.
This strategy not only utilizes the representations from both single branches but also leverages multi-modal fusion representation.
By constructing the positive samples with mutually complementary variables and the negative samples with mutually irrelevant variables, we employ noise contrastive estimation to obtain a proper multi-modal fusion representation via aligning it with video and language branches for better humor detection.

Besides, we also evaluate on the UR-FUNNY\cite{conf/emnlp/HasanRZZTMH19} dataset (from long-form videos with only aligned text, audio and vision features) to further verify the effectiveness of our CVLA.
% detect humor within TED talk punchlines in long-form video scenarios.
% Our CVLA yields superior results on these humor detection datasets.

Overall, our principal contributions are summarized as follows: \\
$\bullet$  We expand the unlabeled data (and keep the labeled data) in DY24h dataset to enhance the robustness of pre-training, yielding a new short-form video dataset, namely DY11k; \\
$\bullet$  We propose a novel \textbf{C}omment-aided \textbf{V}ideo-\textbf{L}anguage \textbf{A}lignment (\textbf{CVLA}) approach via the assistance of comments and data-augmented contrastive pre-training for detecting humor; \\
$\bullet$  Extensive experimental results and analysis on two humor detection datasets (DY11k and UR-FUNNY) demonstrate the effectiveness of CVLA in video humor detection.

%% file: figure/teaser.tex
\begin{figure}[!t]
    \centering
    \includegraphics[width=3.0in]{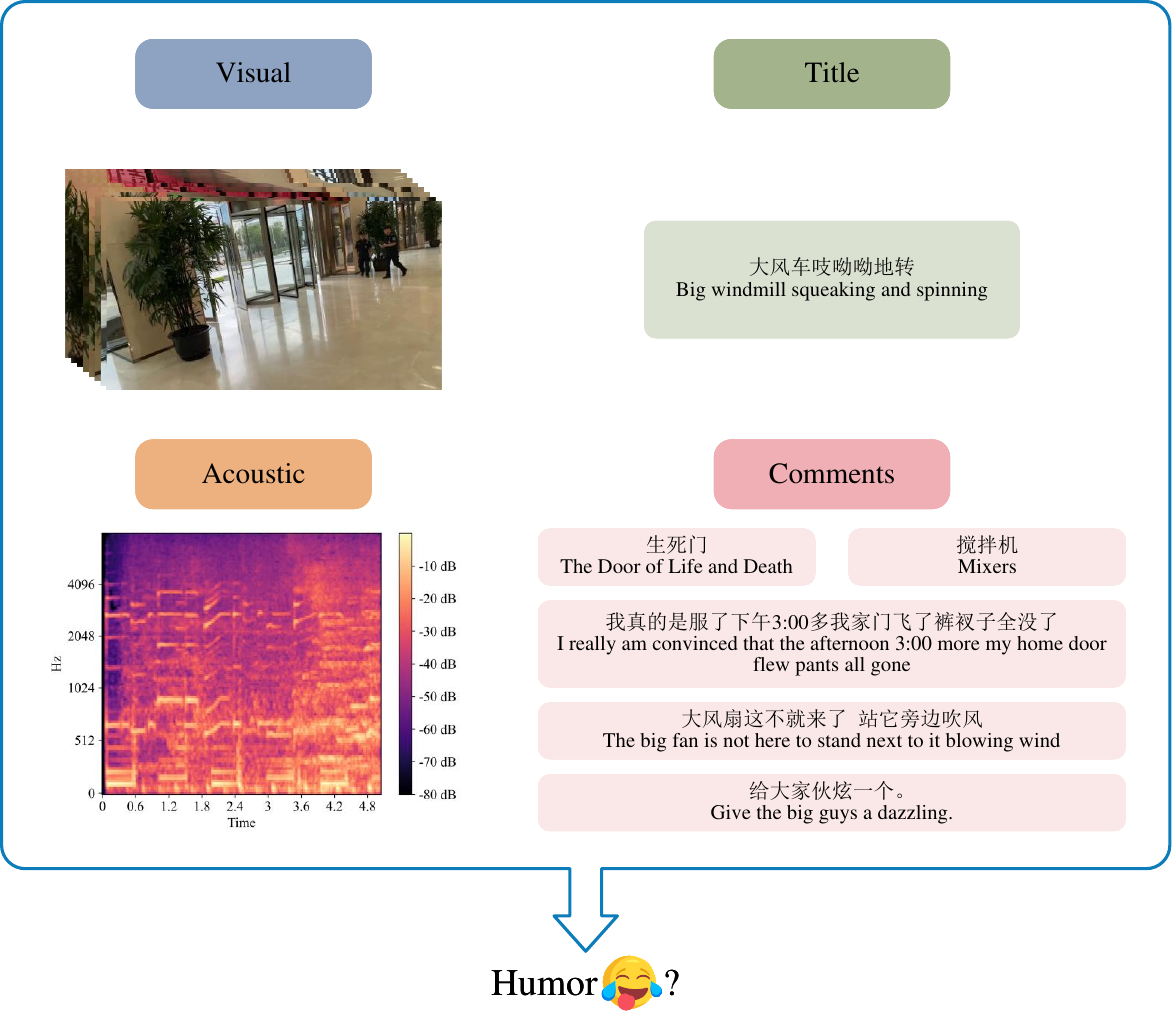}
    \caption{\label{fig:teaser} An example for SVHD.}
\end{figure}

%% file: section/3-related_work.tex
\section{Related Work}

\paragraph{Humor Detection}

Previous studies predominantly concentrate on utilizing deep-learning models for textual humor detection\cite{conf/naacl/ChenS18,conf/acl/BlinovBB19,conf/emnlp/WellerS19,journals/corr/abs-2004-12765,conf/acl/XieLP20,conf/naacl/AoVPA22,conf/wsdm/ChenLLXLC23}.
However, with the growing popularity of multi-modal expressions, multi-modal humor detection becomes an increasingly important area of focus.
\citet{conf/emnlp/HasanRZZTMH19} and \citet{conf/aaai/HasanLR0MMH21} conduct humor detection in the punchline of TED talks, using a multi-modal approach by incorporating the pre-processed features of text (transcriptions), audio, and video.
Additionally, a number of studies\cite{conf/nlpcc/WuLYX21,conf/icmi/ChauhanSM0EBMP21,conf/coling/AlnajjarHTLK22,conf/coling/ChauhanSAEB22} attempt to identify humor labels for each individual utterance in TV show dialogues from diverse cultures mianly towards the long-form videos. Although \citet{conf/ecai/LiuPZSLZ23} also construct a new short-form video dataset (DY24h) with interactive comments for humor detection, 
% and design three heterogeneous proxy tasks for self-supervised learning. 
their dataset is limited and they rely on too many meta data to build the heterogeneous pre-training for video representation learning.

Unlike the above studies, CVLA operates on raw signals across various modal channels eschewing the reliance on pre-extracted features.
We employ a semi-supervised approach CVLA, devising a data-augmented contrastive pre-training strategy that utilizes a substantial corpus of unlabeled short-form videos (even more) for video understanding.
The novel contrastive strategy is applicable to a broad range of tasks in video-language multi-modal analysis.

\paragraph{Video Understanding}

Previous studies in video understanding normally ignore linguistic information (probably not available), thus taking object tracking, action recognition, and object segmentation as the challenges for video understanding\cite{conf/interspeech/RouditchenkoBHC21,NEURIPS2022_3ea31343}.
Although several studies\cite{Wang_2023_CVPR,NEURIPS2022_f8290ccc} start to explore both video and language modalities for better video understanding, they don't have access to the interactive comments on the social aspect.
Besides, CNNs have long been the standard for backbone architectures in this area\cite{conf/iccv/TranBFTP15,conf/cvpr/CarreiraZ17,conf/eccv/XieSHTM18,conf/cvpr/TranWTRLP18,conf/iccv/Feichtenhofer0M19}.
Only Recently, thanks to the great success of Transformer in computer vision, Transformer-based architectures\cite{conf/iclr/DosovitskiyB0WZ21} have attracted much attention in video understanding\cite{conf/icml/BertasiusWT21,conf/cvpr/LiuN0W00022}.
For example, for video emotion detection, \citet{conf/icmi/HanCG0MP21} propose a Transformer-based bimodal fusion approach by modality regularization and gated control mechanism.
Moreover, due to the efficient ability of multi-layer perception (MLP), \citet{conf/mm/SunWL0L22} introduce MLP-based structure from different axes to fuse multi-modality for sentiment analysis.

Different from the above studies, we additionally leverage the potential linguistic knowledge that may be ignored, e.g., comments below a short-form video.
Besides, we propose a contrastive pre-training approach to mine the unlabeled data.
In this way, we can more accurately determine the humor expressed by short-form videos with low resources.

\paragraph{Video-Language Pre-training}

Video-language pre-training has been verified to be effective for various downstream tasks, e.g., text-to-video retrieval and video question-answering\cite{NEURIPS2022_3ea31343,NEURIPS2022_f8290ccc}.
In the literature, general pre-training tasks are normally employed for the downstream tasks, such as masked frame modeling\cite{conf/iccv/SunMV0S19,conf/cvpr/ZhuY20a}, masked modal modeling\cite{conf/acl/XuGHAAFMZ21}, video-language matching\cite{conf/emnlp/LiCCGYL20}, frame ordering modeling\cite{conf/mm/LeiLLHW0TML21,conf/emnlp/LiCCGYL20}.
Although \citet{conf/nips/AkbariYQCCCG21} also propose to align the video and text by multi-modal pre-training, they only aim to enhance the representation of each modality rather than multi-modal representation, then conduct single-modal classification and cross-modal retrieval.

Dissimilar to the above studies, we propose a novel comment-aided video-language alignment approach via contrastive pre-training with a few labeled and numerous unlabeled data.
This approach can align video and language to a consistent semantic space, and simultaneously produce a specific multi-modal representation for short-form video humor detection.

%% file: section/4-method.tex
\input{figure/framework}

\section{Methodology}

In this section, we first give the definition of short-form video detection (SVHD) task.
Then, we present the base framework of our model, including uni-modal and multi-modal encoding, and two-branch hierarchical architecture.
Afterward, we detail our proposed video-language alignment strategy with contrastive pre-training, including cross-modal alignment from a video perspective and language perspective with multi-modality fused representation.
Finally, we adopt the classification network to achieve humor detection.

\subsection{Task Formulation}

Given sample $s_{i}=(v_{i}, a_{i}, t_{i}, c_{i})$, where $v_{i}, a_{i}$ denotes the vision and audio respectively, and $t_{i}, c_{i}$ denotes its accompany textual information including the video title and the interactive comments, we aim to detect whether the video of the sample $s_{i}$ is humorous, by a model $\mathcal{F}(s_{i};\Theta)\mapsto y_{i}$ trained with a few labeled samples and a large number of unlabeled samples.
The annotated sample $s_{i}$ corresponds to a binary label $y_{i}\in\{0, 1\}$, where $y_{i}=1$ indicates humorous, and $y_{i}=0$ means $s_{i}$ not humorous.
\looseness=-1

\subsection{Architecture}
\label{sec:arch}

In this section, we mainly introduce the base modules of our model, excluding the contrastive learning strategy.
Figure~\ref{fig:framework} illustrates the overall framework of our proposed approach for SVHD.
We divide four modalities into two categories—video, which includes vision and audio, and language, encompassing title and comments.
These are then directed into separate branches.
Following this, we implement a two-branch hierarchical network \textbf{C}omment-aided \textbf{V}ideo-\textbf{L}anguage \textbf{A}lignment (\textbf{CVLA}).

\paragraph{Video Encoding}

For video branch, we process the vision and audio features by patchifying:

\textbf{1) Vision:}
Raw video $v_{i}$ can be expressed as $v_{i}\in\mathbb{R}^{48\times3\times224\times224}$ (frames $\times$ channel $\times$ height $\times$ width).
Following previous studies\cite{conf/nips/Feichtenhofer0L22, NEURIPS2022_3ea31343} on video-based understanding, we adopt ViT\cite{conf/iclr/DosovitskiyB0WZ21}-style visual modality embedding.
The patches are flattened and embedded by linear projection, then trainable space-time positional embedding are added to the embedded patches.
The final embedding of vision can be expressed as $e^{v}_{i}\in\mathbb{R}^{\frac{48}{6}\cdot\frac{224}{32}\cdot\frac{224}{32}\times d}$ when patch size is $(6, 32, 32)$ in temporal, height and width dimension respectively, and $d$ denotes the dimension of embedding.

\textbf{2) Audio:}
We extract the log mel-spectrogram $a_{i}\in\mathbb{R}^{\frac{o\cdot8000}{512}\times128}$ for audio at $8k$-Hz with a total duration of $o$ seconds and $512$ samples between successive audio frames (hop length).
Then, we treat the $2$-dimension audio mel-spectrogram as an image, divide it into patches.
Similar to vision, the patches are flattened and embedded by a linear projection, then add the trainable positional embedding.
The final embedding of audio can be expressed as $e^{a}_{i}\in\mathbb{R}^{\frac{o\cdot8000/512}{16}\cdot\frac{128}{16}\times d}$ when patch size is $(16, 16)$ in time and frequency axis respectively.

On the above basis, we treat both visual and auditory data as components of the video branch.
Therefore, we input them as a sequence into an multi-layer Transformer\cite{conf/nips/VaswaniSPUJGKP17} encoder, where they completely interact with each other via self-attention mechanism:
\begin{gather}
V_{i}=\textrm{VE}(\textrm{[CLS]}; e^{v}_{i}; \textrm{[CLS]}; e^{a}_{i})
\end{gather}
where $\textrm{VE}$ indicates the video encoder.
$X; Y$ denotes concatenate operation in sequence dimension between $X$ and $Y$, and we denote the output $V^e_{i}$ (from $\textrm{[CLS]}$ token in $V_{i}$) to represent the video features of the $i$-th sample.

\paragraph{Language Encoding}

For language branch, we tokenize title and comments then encode:

\textbf{1) Title:}
Each token (e.g., the $t$-th token)in the $i$-th title is represented as the sum of word embedding $t^w_{i,k}$, position embedding $t^p_{i,k}$.
Formally: $t_{i,k} = t^w_{i,k}+t^p_{i,k}$, where $t_{i,k}$ denotes the $k$-th token representation input into our model.
Thus, $t_{i}$ indicates the whole title sequence of $i$-th sample.

\textbf{2) Comment:}
Similarly, each token (e.g., the $k$-th token) in the $j$-th comment of the $i$-th sample is represented as the sum of word embedding $c^w_{i,j,k}$, position embedding $c^p_{i,j,k}$.
Formally: $c_{i,j,k} = c^w_{i,j,k}+c^p_{i,j,k}$, where $c_{i,j,k}$ denotes the $k$-th token in the $j$-th comment of the $i$-th sample.
Thus, the whole comment sequence can be represented as $\{c_{i,1}, \cdots, c_{ij}\}$. Note that every instance contains up to $10$ comments.

On the above basis, we treat both title and comments data as components of the language branch.
Therefore, we cast the title and comments as one document but distinguish them with the $\textrm{[SEP]}$ token, and add the $\textrm{[CLS]}$ token at the beginning.
Then, we employ the pre-trained language model BERT\cite{conf/naacl/DevlinCLT19} to encode them:
\begin{gather}
L_{i}=\textrm{LE}(\textrm{[CLS]}; {t}_{i}; \textrm{[SEP]}; {c}_{i,1}; \cdots; {c}_{i,j}; \textrm{[SEP]})
\end{gather}
where $\textrm{LE}$ indicates the language encoder BERT that can be easily replaced with Large Language Model (LLM).
We take the output $L^e_{i}$ (from $\textrm{[CLS]}$ token in $L_{i}$) to represent the language features of the $i$-th sample.

\paragraph{Multi-modal Fusion}

To make multi-modal data completely interact with each other and produce a unified multi-modal fusion representation, we employ self-attention mechanism\cite{conf/icassp/RajanBC22} by the Transformer-based multi-modal encoder ($\textrm{MME}$) that can be easily replaced with Large Vision-Language Model (LVLM).
Formally:
\begin{gather}
M_{i}=\textrm{MME}(V_{i}; L_{i})
\end{gather}
where we denote the output $M^e_{i}$ (from the first token $\textrm{[CLS]}$ in $M_{i}$) to represent the multi-modal features of $i$-th sample.

\subsection{Data-augmented Contrastive Pre-training}
\label{sec:pre}

To effectively align the video and language with multi-modal representation, we attempt contrastive learning, which can also utilze the large-scale unlabeled data at the pre-training stage.
In the literature, data augmentation is a common strategy for contrastive learning to construct positive pairs, such as SimCLR\cite{conf/icml/ChenK0H20} in computer vision (CV) and SimCSE\cite{conf/emnlp/GaoYC21} in natural language processing (NLP).

\paragraph{Data Augmentation}

For the video branch, to adapt our task, we randomly apply three of the \textit{Erase} ($10\sim20\%$ area), \textit{Color jitter}, \textit{2D affine}, \textit{Color drop}, \textit{Gaussian blur}, and \textit{Gaussian noise} following \citet{conf/icml/ChenK0H20} to the visual modality with a probability of $33.\dot{3}\%$.
Besides, we randomly remove $10\sim20\%$ of the mel-spectrogram for audio.
To this end, we feed original patches of vision and audio into the video encoder ($\textrm{VE}$) to get $V_{i}$.
Meanwhile, we also feed augmented vision and audio patches data from the same sample into $\textrm{VE}$ module to obtain $\tilde{V}_{i}$, then pair both $\textrm{[CLS]}$ tokens of them as $(V^{e}_{i},\tilde{V}^{e}_{i})$.

For the language branch, we feed the title and comments of a sample into the language encoder ($\textrm{LE}$) twice to obtain the different language sequence representation (due to dropout probability), i.e., $L_{i}$ and $\tilde{L}_{i}$, respectively.
Then, we pair both $\textrm{[CLS]}$ token of them as $(L^{e}_{i},\tilde{L}^{e}_{i})$.

Finally, to obtain the proper representation for better SVHD, we need to involve fused multi-modal representation in contrastive pre-training.
Therefore, we adopt the multi-modal encoder ($\textrm{MME}$) to yield two kinds of fused multi-modal representations $\tilde{M}^{e}_{i}$ and $M^{e}_{i}$ from $(\tilde{V}^{e}_{i},L^{e}_{i})$ and $(V^{e}_{i},\tilde{L}^{e}_{i})$ as input, respectively.
From video perspective, $\tilde{M}^{e}_{i}$ can be considered as the additional knowledge towards original video representation $V^{e}_{i}$ in that the used input $(\tilde{V}^{e}_{i}$ and $L^{e}_{i})$ of $\tilde{M}^{e}_{i}$ are very different from $V^{e}_{i}$.
Similarly, from language perspective, $M^{e}_{i}$ can be considered as the additional knowledge towards original language representation $L^{e}_{i}$.

\paragraph{Contrastive Tuples Construction}

For the positive tuple from the video perspective, we believe that the original video representation $V^{e}_{i}$ should accept the help of other modalities (i.e., $L^{e}_{i}$ and $\tilde{L}^{e}_{i}$), as well as the fused multi-modal representation $\tilde{M}^{e}_{i}$ with additional knowledge.
Note that although the original video and language modalities may exist gap, after modality encoding, we expect them to achieve semantic alignment.
Therefore, we can obtain the positive tuple ($V^{e}_{i}, L^{e}_{i}, \tilde{L}^{e}_{i}, \tilde{M}^{e}_{i}$), then reduce the distance of latent space among them.
Through subsequent contrastive learning, this also can be considered as to augment the video and multi-modal representation for our humor detection.
Regarding the negative tuples, in a batch of the training samples, we take the contrary guidance from other samples, i.e., ($V^{e}_{i}, L^{e}_{j}, \tilde{L}^{e}_{j}, \tilde{M}^{e}_{j}$), where $j\neq i$.
Similarly, from the language perspective, we can obtain the positive tuple ($L^{e}_{i}, V^{e}_{i}, \tilde{V}^{e}_{i}, M^{e}_{i}$) to augment the language and multi-modal representations.
Besides, the negative tuple is defined as ($L^{e}_{i}, V^{e}_{j}, \tilde{V}^{e}_{j}, M^{e}_{j}$) where $j\neq i$ to keep distance among them.

\paragraph{Contrastive Pre-Training}

For the contrastive learning among the multiple variables, we leverage the noise contrastive estimation (NCE) loss\cite{Huang_2023_CVPR} within a batch $\mathcal{B}$.

\input{figure/dataset}

From the \textit{video perspective}, we compute the NCE loss of the positive and negative tuples in a batch as follows:
\begin{gather}
\mathcal{L}_{v}=-\sum^{\mathcal{B}}_{i=1}[g(V^{e}_{i},L^{e}_{i})+g(V^{e}_{i},\tilde{L}^{e}_{i})+g(V^{e}_{i},\tilde{M}^{e}_{i})]  \label{eq:loss_v} \\
g(p,q)=\log{\frac{f(p,q)}{f(p,q)+Z}} \\
Z=\sum^{\mathcal{B}}_{j=1,j\neq i}[f(V^{e}_{i},L^{e}_{j})+f(V^{e}_{i},\tilde{L}^{e}_{j})+f(V^{e}_{i},\tilde{M}^{e}_{j})] \label{eq:eq_z} \\
f(p,q)=e^{\textrm{sim}(p,q)/\tau}
\end{gather}
where $p$ and $q$ are the formal parameters, which can accept the variables in positive and negative tuples.
$\tau$ denotes a trainable temperature scalar parameter, and $\textrm{sim}(\cdot,\cdot)$ is the dot-product similarity function to measure the degree of alignment between different representations.
To avoid the sub-optimal situation (e. g., $L^{e}_{i}$ may be closer to $V^{e}_{i}$ while $\tilde{L}^{e}_{i}$ and $\tilde{M}^{e}_{i}$ become away from $V^{e}_{i}$.), we further narrow the semantic distance between the video and the complementary representations of language and multi-modal fusion, in which the semantic distance are highlighted by normalization:
\begin{gather}
\mathcal{L}_{v^{\prime}}=-\sum^{\mathcal{B}}_{i=1}[\varphi(L^{e},V^{e},i)+\varphi(\tilde{L}^{e},V^{e},i)+\varphi(\tilde{M}^{e},V^{e},i)] \label{eq:loss_vp} \\
\varphi(p,q, i)=\log{\frac{f(p_{i},q_{i})}{\sum^{B}_{j=1}f(p_{i},q_{j})}}
\end{gather}
The two losses in~\ref{eq:loss_v} and~\ref{eq:loss_vp} are summed up to give the learning objective of alignment: $\mathcal{L}_{V}=\mathcal{L}_{v}+\mathcal{L}_{v^{\prime}}$ from the video perspective.

Similarily, from the \textit{language perspective}, we define the alignment objective $\mathcal{L}_{L}=\mathcal{L}_{l}+\mathcal{L}_{l^{\prime}}$.
Then, we obtain the final learning objective in pre-training:
\begin{gather}
\mathcal{L}=\mathcal{L}_{V}+\mathcal{L}_{L}+\lambda||\Theta||^{2}
\end{gather}
where $\Theta$ denotes all trainable parameters of the model $\mathcal{F}$, $\lambda$ represents the coefficient of $L_{2}$-regularization.

\subsection{Humor Detection}

In the fine-tuning stage, we do not need to use the procedure of data augmentation, and we adopt multi-modal hybrid representation $M^{e}_{i}$ to infer the humor status: $\hat{y}_{i}=\textrm{Softmax}(\textrm{MLP}(M^{e}_{i}))$.
Here, cross-entropy is adopted for the learning objective:
\begin{gather}
\mathcal{J}=-\sum^{N}_{i=1}y_{i}\log{\hat{y_{i}}}+\lambda||\Theta||^{2}
\end{gather}
where $N$ denotes training set size, $\Theta$ denotes all trainable parameters of the model $\mathcal{F}$, $\lambda$ represents the coefficient of $L_{2}$-regularization.

%% file: figure/framework.tex
\begin{figure*}[!t]
    \centering
    \includegraphics[width=6.0in]{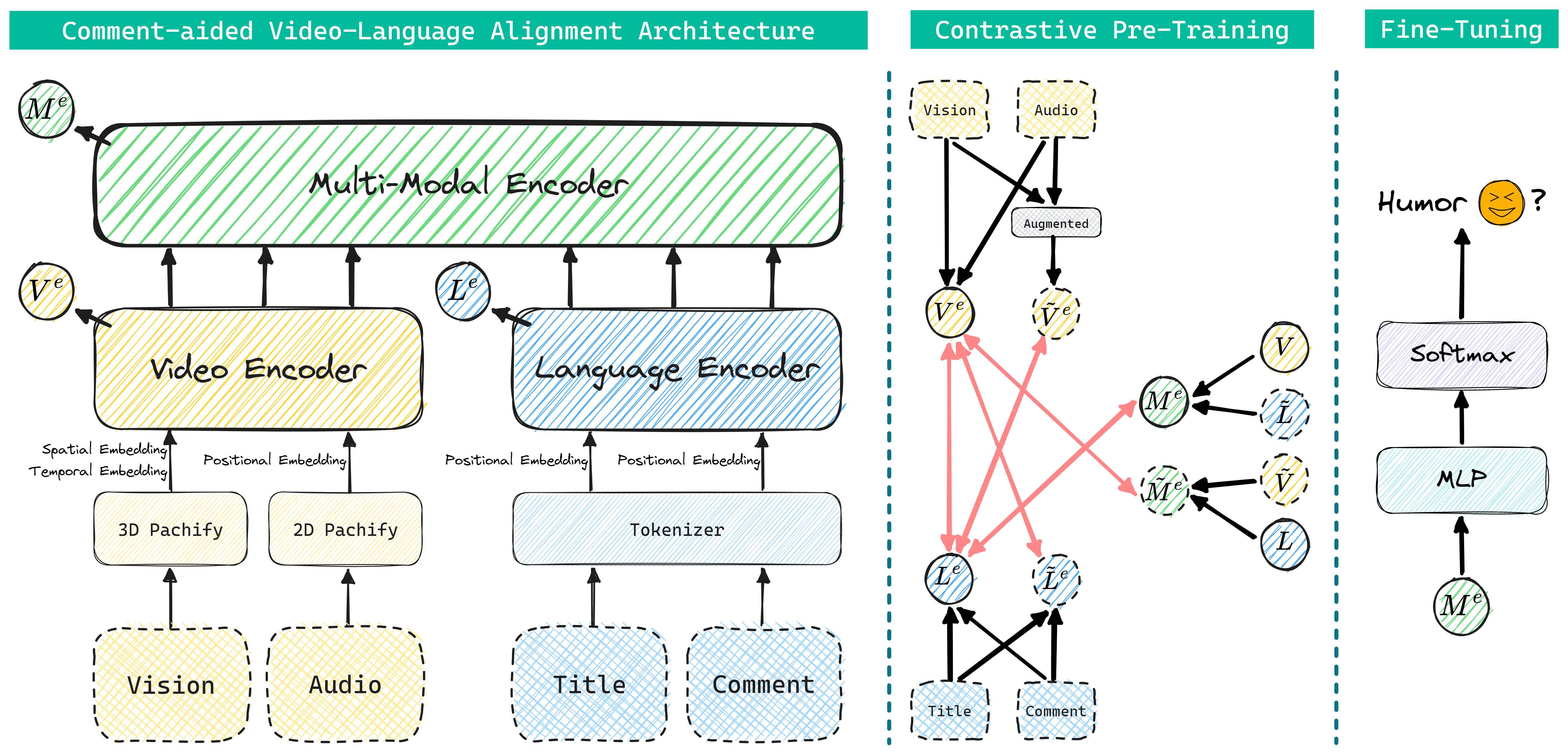}
    \caption{\label{fig:framework} The overall framework of our proposed CVLA for short-form video humor detection.}
\end{figure*}

%% file: figure/dataset.tex
\begin{figure*}[!t]
  \centering
  \subfigure[\label{fig:duration} Duration of videos distribution.]{
    \includegraphics[width=0.3\linewidth,trim=5 0 5 15,clip]{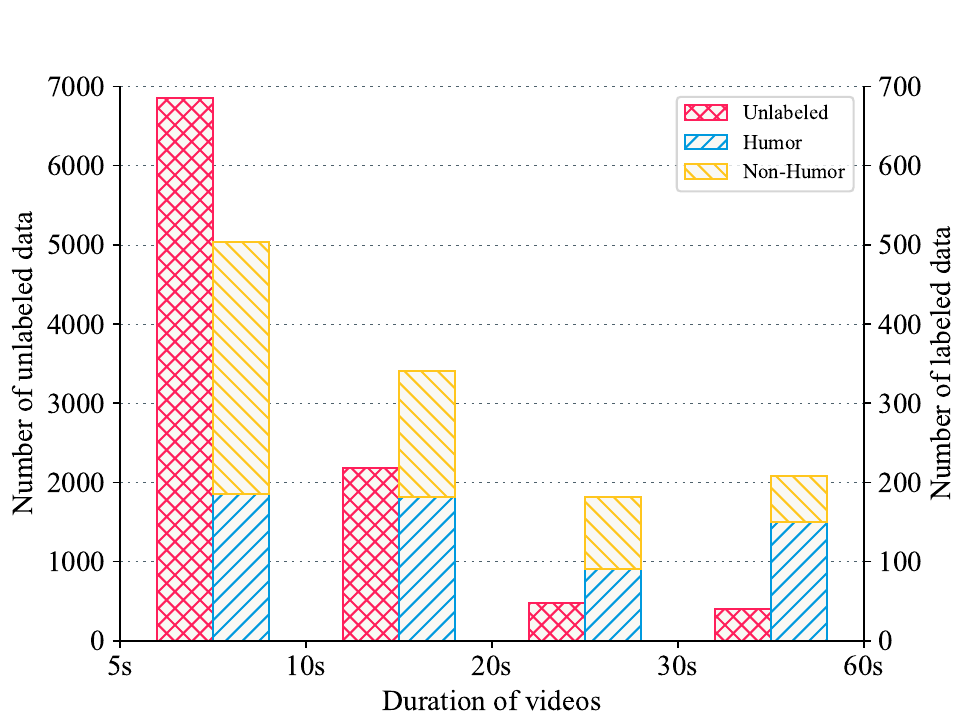}
  }
  \hfill
  \subfigure[\label{fig:like} Number of likes distribution.]{
    \includegraphics[width=0.3\linewidth,trim=5 0 5 15,clip]{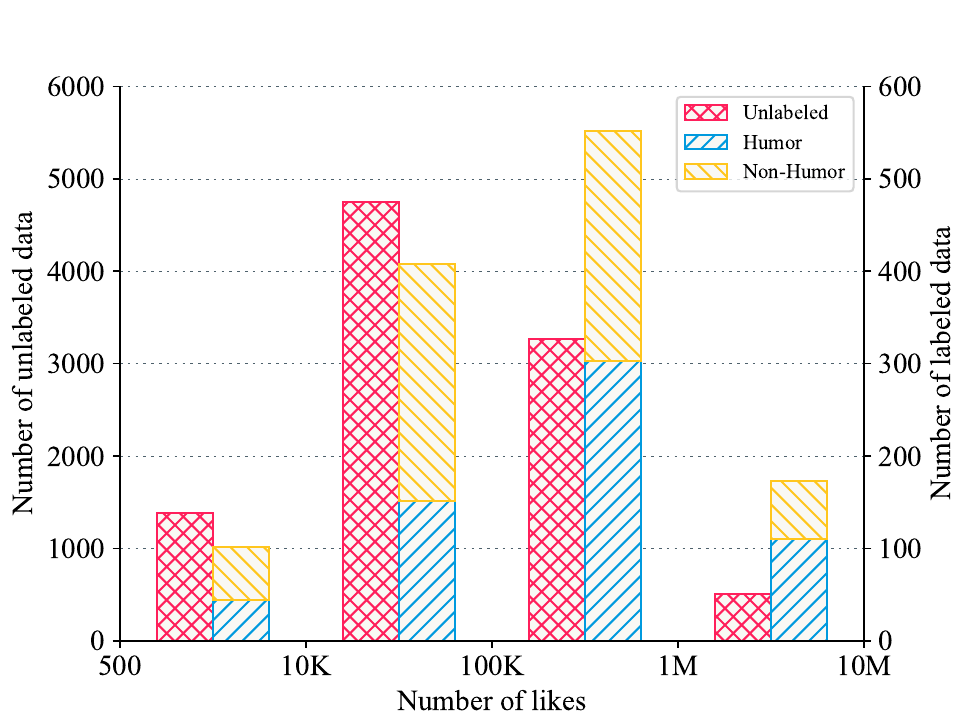}
  }
  \hfill
  \subfigure[\label{fig:comment} Number of comments distribution.]{
    \includegraphics[width=0.3\linewidth,trim=5 0 5 15,clip]{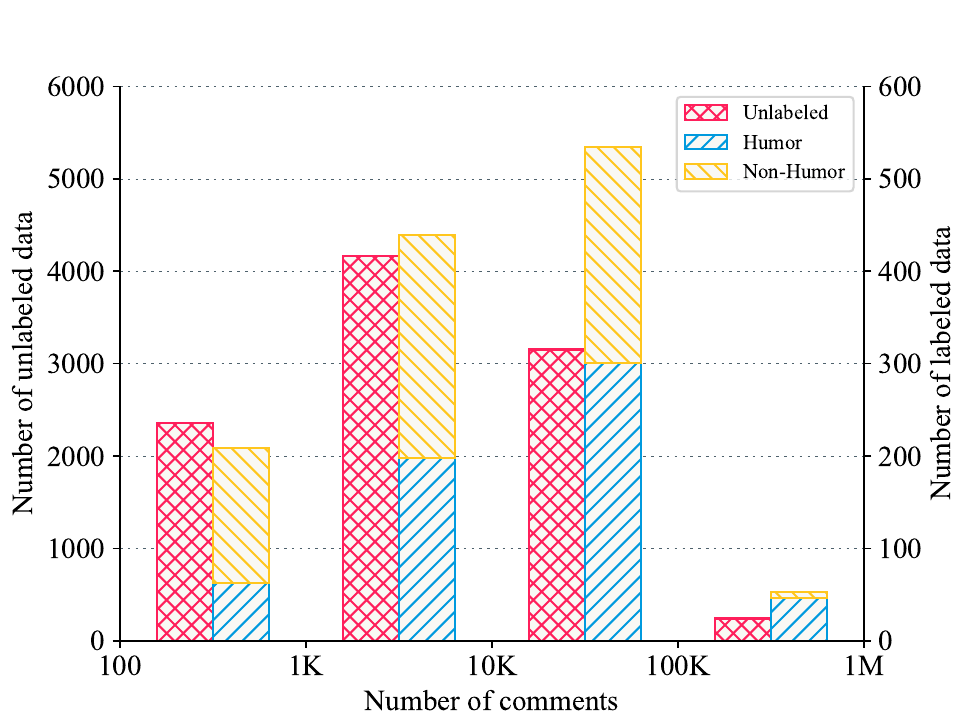}
  }
  \caption{\label{fig:dataset} The statistics of our expanded dataset DY11K.}
\end{figure*}

%% file: section/5-experiments.tex
\section{Experimentation}

\subsection{Data Settings}
\label{sec:dataset}

In our experiments, we utilized the DY24h\cite{conf/ecai/LiuPZSLZ23} dataset, supplemented with additional unlabeled data for pre-training, named DY11k.
Further, we conducted experiments on the UR-FUNNY\cite{conf/emnlp/HasanRZZTMH19} dataset to detect humor labels in TED talk punchlines.

\paragraph{Data Expansion}

Adhering to the data collection protocols established for DY24h\cite{conf/ecai/LiuPZSLZ23}, we have amassed an additional $3060$ unlabeled short-form video samples, each encompassing four modalities: vision, audio, title, and comments.
These samples are incorporated into the unlabeled data part of DY24h for the pre-training stage.
Following this inclusion, the expanded dataset, now known as DY11k, consists of a total of $11150$ samples, amounting to $35.38$ hours.
Of these, $9915$ samples ($29.37$ hours) without annotation are used to conduct self-supervised pre-training, while the remaining $1235$ samples ($6.01$ hours) are manually annotated to perform supervised learning and testing for humor detection same as DY24h.
The statistic summary can refer to Figure~\ref{fig:dataset}.

\input{table/dataset}

\paragraph{Data Split}

For DY11k, we randomly allocate the $1235$ labeled samples into training, development, and test sets, ensuring a balanced distribution across labels, facilitated by five distinct random seeds (the training of the model employs the corresponding random seed used for data segmentation).
For UR-FUNNY, we segment the $6334$ available samples with raw videos into unlabeled and labeled groups.
The $4634$ unlabeled samples is utilized for pre-training stage without regard to its labels.
Similar to DY11k, we randomly allocate the other $1700$ labeled samples into training, development, and test sets.
Table~\ref{tab:dataset} delineates the specifics of the data division for both humor detection datasets.

\input{table/main}
\input{table/ur_funny}

\subsection{Implementation Details}

We uniform sampling $48$ frames for video and resize frames to a spatial size of $224\times224$, and patch size is $(6, 32, 32)$ for it.
We use mel-spectrogram of librosa\cite{conf/scipy/McFeeRLEMBN15} with $8$-kHz sampling rate, $512$ hop length and $128$ mel bands, and patch size is $(16, 16)$ for it.
The maximum length of language (title and comments) token sequence is $16$ for each sentence, and we discard all the \textit{\#Funny} and \textit{\#Funny Video} hashtags.
AdamW\cite{conf/iclr/LoshchilovH19} is utilized as the optimizer and the learning rate is $0.00001$ for both pre-training and fine-tuning.
The coefficient of $L_{2}$-regularization $\lambda$ is set to $0.01$ for all optimizer.
The dimension of all embedding and encoding in our model is $512$.
The number of heads in the Transformer\cite{conf/nips/VaswaniSPUJGKP17} encoder is set to $12$, and the encoder is set to $8$ layers. The dropout rate is set to $0.1$.

Mean and standard deviation are calculated based on five runs (remove the top and bottom) with different random seeds to avoid unstable extreme values.
Multiple dataset partitions and model training give a more robust measure of performance and a better estimate of the standard.
We pre-train our model with batch size $6$ and epoch $10$ on a single Nvidia RTX $3090$ ($24$GB VRAM) under three hours via the Pytorch\cite{conf/nips/PaszkeGMLBCKLGA19} toolkit.

\subsection{Baselines}

% To put our results in perspective, we compare to a number of baselines: \\
$\bullet$ \textbf{TVLT}\cite{NEURIPS2022_3ea31343}: a state-of-the-art (SOTA) multi-modal approach for video understanding (e.g., image retrieval, video retrieval, and multi-modal sentiment analysis) with video-audio pre-training on datasets HT100M\cite{conf/iccv/MiechZATLS19} and YTT180M\cite{conf/nips/ZellersLHYPCFC21}.
However, since the model is particularly designed for vision and audio modality, we can not directly input language modality; \\
$\bullet$ \textbf{BERT}\cite{conf/naacl/DevlinCLT19} and \textbf{DeBERTa}\cite{conf/iclr/HeLGC21}: the SOTA approaches for natural language understanding with large-scale pre-training on multiple corpus.
We conduct humor detection as text classification task; \\
$\bullet$ \textbf{LF-VILA}\cite{NEURIPS2022_f8290ccc}: the SOTA approach for long-form (similar to short-form setting in DY11k) video understanding (e.g., cross-modal retrieval and long-form video classification) with video-language pre-training on their dataset LF-VILA-8M\cite{NEURIPS2022_f8290ccc}.
We adapt the title and comments in DY11k to the language modality of this model LF-VILA, then conduct humor detection.
However, since this model is particularly designed for vision and language modality, we can not directly input audio; \\
$\bullet$ \textbf{C-MFN}\cite{conf/emnlp/HasanRZZTMH19}: the baseline approach for UR-FUNNY\cite{conf/emnlp/HasanRZZTMH19} dataset, that designed for the detection of punchline humor in TED talks, accompanied by synchronous transcriptions in long-form videos; \\
$\bullet$ \textbf{BBFN}\cite{conf/icmi/HanCG0MP21} and \textbf{CubeMLP}\cite{conf/mm/SunWL0L22}: the SOTA approaches for emotion detection of videos without multi-modal pre-training; \\
$\bullet$ \textbf{VATT}\cite{conf/nips/AkbariYQCCCG21}: the SOTA approach in multi-modal video understanding (e.g., audio event classification, image classification, and text-to-video retrieval) with video-audio-language pre-training; \\
$\bullet$ \textbf{M3GAT}\cite{journals/tois/ZhangJWZZLHJSQ24}: the SOTA approach for multi-modal dialogue sentiment analysis and emotion recognition; \\
$\bullet$ \textbf{CMHP}\cite{conf/ecai/LiuPZSLZ23}: the SOTA approach for short-form video detection with multi-modal heterogeneous pre-training and the assistance of interactive comments.

Note that for a fair comparison, we adapt the title and comments in DY11k to the language modality (dialogue for M3GAT) of these model.
We also implement these pre-training approach on DY11k and UR-FUNNY, then conduct humor detection.

\input{table/effect}

\subsection{Main Results}

Tables~\ref{tab:result} and~\ref{tab:ur_funny} present the comparative performance of the CVLA against SOTA baselines on two humor detection datasets.

Experimental results of Tables~\ref{tab:result} and~\ref{tab:ur_funny} reveals that:
1) CVLA surpasses all baselines obviously, demonstrating the advantage of our architecture even without pre-training on DY11k. This is attributed to effective modality encoding and the multi-modal  of video and language.
Crucially, the video-language alignment facilitated by contrastive pre-training is instrumental, generating a nuanced multi-modal representation for humor detection.
Moreover, existing approaches such as C-MFN, BBFN rely on the feature provided by UR-FUNNY dataset, while CMHP relies on the feature extracted by ResNet\cite{conf/cvpr/HeZRS16}, which may result in the omission of important information.
In contrast, our CVLA differs from the above since it operates on raw signals across various modal channels, the trait enhances performance and also streamlines the training and inference process.
2) Baselines lacking pre-training exhibit markedly lower performance compared to their pre-trained counterparts, with the exception of VATT.
This discrepancy suggests that BBFN, CubeMLP, M3GAT, and C-MFN do not fully leverage the available unlabeled data, potentially overlooking critical information for humor detection.
VATT, notably, completely ignores the explicit multi-modal fusion and multi-modal representation enhancement during its pre-training stage.
In terms of standard deviation, our CVLA is steadily superior to than BBFN, despite the performance are similar on UR-FUNNY.
Although the accuracy of LF-VILA and CMHP is similar to that of C-MFN, there is a significant gap in their F1 values.
3) The variation in performance of baselines pre-trained on different datasets indicates inconsistent task adaptation among existing methods.
For instance, TVLT, pre-trained on a combined dataset of HT100M and YTT180M comprising approximately $280$ million clips, shows improved results over those achieved using our DY11k, which contains roughly $10k$ unlabeled samples.
Conversely, performance of LF-VILA on its LF-VILA-8M dataset is slightly inferior to that on DY11k.
Therefore, existing multi-modal pre-training does not always result in stable task adaptability and may instead enhance only individual modality representations.

These findings substantiate the efficacy of our approach CVLA in addressing the distinctive challenges posed by short-form video humor detection and suggest its applicability to long-form video contexts as well.

\subsection{Analysis and Discussion}

\paragraph{The effect of different modality combinations}

Table~\ref{tab:effect_modal} shows the performance comparison of our CVLA using different modality combinations without and with our contrastive pre-training.
From this table, we can observe that:
1) As long as using contrastive pre-training, CVLA can perform much better than not using pre-training, regardless of the scenario on different modality combinations.
This demonstrates the effectiveness of our proposed video-language contrastive pre-training strategy;
2) Without pre-training, CVLA on a bimodal scenario of T+C achieves the best performance due to the BERT's excellent language understanding, even better than using three or four modalities, as it is impractical for the video branch to understand the video content with only $100$ labeled training samples and without any pre-trained visual backbone due to the computing resource constraint.
This suggests that although our model without pre-training has extracted features of each modality well and performed nicely in many scenarios, it does not fully exploit the advantages of each modality and the complementarity among different modalities more than two;
Therefore, we attempt to introduce a video-language alignment approach via contrastive pre-training, simultaneously producing a hybrid multi-modal representation for better SVHD.
The effectiveness of this attempt is verified by the results of CVLA with pre-training.
For example, with our pre-training, CVLA based on V+T+C and A+T+C outperform T+C, in which V and A are well utilized based on T+C.

\input{figure/curve}

\paragraph{The indicator trend with and without pre-training}

Figure~\ref{fig:curve} shows the loss-decreasing trend during pre-training (PT) and the accuracy-increasing trend of development set during fine-tuning after PT and validation without PT by our CVLA.
As the course of $10$ epochs in pre-training, loss quickly decreases from the initial $16$ to approximately $3$. This suggests that this phase allows the model to have a solid grasp of the video-language alignment and multi-modal fusion supplied by large-scale short-form videos.
Besides, regarding the accuracy-increasing trend of development set, we can guess that the pre-trained model has captured a large amount of unlabeled generic knowledge, e.g., video-language semantic aligned representation.
Then, by fine-tuning, the model can produce better specific multi-modal representations for humor detection, compared to not using pre-training.

\input{table/scale}
\input{figure/attn}

\paragraph{The effect of pre-training scale}

Table~\ref{tab:effect_scale} displays the performance variation resulting from our CVLA using pre-trained data of different sizes.
From this table, we can see that using too less pre-training data (e.g., 3K) may lead to worse performance.
This is mainly because our contrastive pre-training requires an appropriate data scale to take advantage of its benefits for SVHD.
Besides, from the overall trend, our CVLA can consistently improve the performance by feeding more unlabeled data.
However, due to the limited computational resources, we did not explore more data, which becomes our future work.

\paragraph{The ablation study}

Table~\ref{tab:effect_contrast} shows the performance comparison of different ablated approaches from our CVLA.
The approaches from top to bottom are our full model (CVLA), completely removing the MME block and concatenating $V^{e}_{i}$ and $L^{e}_{i}$ for humor detection (w/o MME), removing of multi-modal representation learning during pre-training (w/o $M^{e}$ ), removing the loss $\mathcal{L}_{V}$ (w/o $\mathcal{L}_{V}$), and removing the loss $\mathcal{L}_{L}$ (w/o $\mathcal{L}_{L}$), removing $\mathcal{L}_{v^{\prime}}$ and $\mathcal{L}_{l^{\prime}}$ (w/o $\mathcal{L}_{v^{\prime}}+\mathcal{L}_{l^{\prime}}$).
From this table, we can find that the additional losses ($\mathcal{L}_{v^{\prime}}+\mathcal{L}_{l^{\prime}}$) avoiding the sub-optimal problem bring in the minimal impact, but it should not be ignored.
In general, removing any of the blocks leads to obvious performance degradation.
This shows that each block of our CVLA has its own necessity.

\input{table/ablation}
\input{figure/tsne}

\paragraph{The visualization in multi-modal encoder}

First, we provide the visualization of self-attention in the multi-modal encoder (MME) using average attention score of  all attention heads, layers, and test samples without and with our proposed contrastive pre-training strategy, as illustrated in Figure~\ref{fig:attn}.
From this figure, we can observe that without pre-training, the video and language branches do not have any mutual interactions, i.e. it is impossible to semantically align them.
However, after using pre-training, the video clearly starts to pay attention to the language features.
In other words, the video representation tries to move closer to the language.

Then, we illustrate the t-SNE visualization of our multi-modal fusion representation $M^{e}$ without and with our pre-training strategy, as shown in Figure~\ref{fig:tsne}.
From this figure, we can observe that the distribution after pre-training becomes more uniform and the distinction between humor labels is clearer, compared to the distribution without pre-training.
This suggests our CVLA can learn a proper representation for better SVHD by contrastive pre-training.

%% file: table/dataset.tex
\begin{table}[!t]
    \small
    \centering
    \caption{\label{tab:dataset} The data split of two humor detection datasets. DY11k expanded frrom DY24h \cite{conf/ecai/LiuPZSLZ23}, UR-FUNNY \cite{conf/emnlp/HasanRZZTMH19}.}
    \begin{tabular}{lcccc}
        \toprule
        \multirow{2}{*}{\textbf{Dataset}} & \multirow{2}{*}{\textbf{Unlabeled}} & \multicolumn{3}{c}{\textbf{Labeled}} \\
        \cmidrule{3-5}
        & & \textbf{Train} & \textbf{Dev} & \textbf{Test} \\
        \midrule
        DY11k & 9915 & 100 & 100 & 1035 \\
        UR-FUNNY & 4634 & 600 & 300 & 800 \\
        \bottomrule
    \end{tabular}
\end{table}

%% file: table/main.tex
\begin{table*}[!t]
    \small
    \centering
    \caption{\label{tab:result} Performance comparison on DY11k dataset. V, A, T and C indicate different modalities of visual, acoustic, title, and comments accordingly. HT100M \cite{conf/iccv/MiechZATLS19}, YTT180M \cite{conf/nips/ZellersLHYPCFC21}, LF-VILA-8M \cite{NEURIPS2022_f8290ccc}. $\uparrow$ denotes the datasets Books and Wiki in the lastrow.}
    \begin{tabular}{clcccccccc}
        \toprule
        \multirow{2}{*}{\textbf{Modality}} & \multicolumn{1}{c}{\multirow{2}{*}{\textbf{Approach}}} & \multirow{2}{*}{\textbf{Pre-Training}} & \multirow{2}{*}{\textbf{Acc}} & \multicolumn{3}{c}{\textbf{Macro}} & \multicolumn{3}{c}{\textbf{Weighted}} \\
        \cmidrule(lr){5-7} \cmidrule(lr){8-10}
         &  &  &  & \textbf{Pre} & \textbf{Rec} & \textbf{F1} & \textbf{Pre} & \textbf{Rec} & \textbf{F1} \\
        \midrule
        \multirow{2}{*}{V+A} & \multirow{2}{*}{TVLT \cite{NEURIPS2022_3ea31343}} & DY11k & \ms{68.12}{0.44} & \ms{69.35}{0.94} & \ms{68.39}{0.59} & \ms{67.80}{0.47} & \ms{69.49}{1.04} & \ms{68.12}{0.44} & \ms{67.72}{0.46} \\
         &  & HT100M+YTT180M & \ms{75.62}{0.92} & \ms{76.08}{0.81} & \ms{75.60}{0.93} & \ms{75.49}{1.00} & \ms{76.08}{0.81} & \ms{75.62}{0.92} & \ms{75.50}{0.99} \\
        \cmidrule(lr){1-10}
        \multirow{2}{*}{T+C} & BERT \cite{conf/naacl/DevlinCLT19} & Books+Wiki & \ms{74.33}{1.49} & \ms{74.59}{1.48} & \ms{74.40}{1.47} & \ms{74.30}{1.49} & \ms{74.64}{1.48} & \ms{74.33}{1.49} & \ms{74.28}{1.49} \\
         & {DeBERTa \cite{conf/iclr/HeLGC21}} & $\uparrow$+WebText+Stories & \ms{70.79}{0.87} & \ms{71.03}{0.99} & \ms{70.88}{0.89} & \ms{70.75}{0.84} & \ms{71.09}{1.00} & \ms{70.79}{0.87} & \ms{70.74}{0.83} \\
        \cmidrule(lr){1-10}
        \multirow{2}{*}{V+T+C} & \multirow{2}{*}{LF-VILA \cite{NEURIPS2022_f8290ccc}} & DY11k & \ms{76.23}{0.96} & \ms{77.02}{1.92} & \ms{76.42}{1.12} & \ms{76.14}{0.85} & \ms{77.15}{2.04} & \ms{76.23}{0.96} & \ms{76.11}{0.82} \\
         &  & LF-VILA-8M & \ms{75.14}{2.02} & \ms{75.46}{1.75} & \ms{75.23}{1.93} & \ms{75.09}{2.06} & \ms{75.52}{1.70} & \ms{75.14}{2.02} & \ms{75.07}{2.08} \\
        \cmidrule(lr){1-10}
         & BBFN \cite{conf/icmi/HanCG0MP21} & {\color{MyRed}\ding{55}} & \ms{74.40}{0.16} & \ms{74.80}{0.20} & \ms{74.46}{0.19} & \ms{74.31}{0.26} & \ms{74.86}{0.17} & \ms{74.40}{0.16} & \ms{74.31}{0.26} \\
         & VATT \cite{conf/nips/AkbariYQCCCG21} & DY11k & \ms{71.98}{3.01} & \ms{72.54}{2.91} & \ms{72.06}{3.09} & \ms{71.82}{3.14} & \ms{72.60}{2.98} & \ms{71.98}{3.01} & \ms{71.81}{3.13} \\
         & CubeMLP \cite{conf/mm/SunWL0L22} & {\color{MyRed}\ding{55}} & \ms{73.66}{3.25} & \ms{74.10}{3.03} & \ms{73.71}{3.32} & \ms{73.53}{3.40} & \ms{74.15}{3.08} & \ms{73.66}{3.25} & \ms{73.52}{3.39} \\
         & M3GAT \cite{journals/tois/ZhangJWZZLHJSQ24} & {\color{MyRed}\ding{55}} & \ms{67.23}{2.44} & \ms{67.28}{2.51} & \ms{67.20}{2.55} & \ms{67.12}{2.51} & \ms{67.32}{2.56} & \ms{67.16}{2.48} & \ms{67.12}{2.51} \\
         & \multirow{1}{*}{CMHP \cite{conf/ecai/LiuPZSLZ23}} 
         & DY11k & \ms{71.48}{1.93} & \ms{72.60}{1.64} & \ms{71.51}{1.73} & \ms{71.12}{2.02} & \ms{72.63}{1.54} & \ms{71.48}{1.93} & \ms{71.12}{2.08} \\
        \cmidrule(lr){2-10}
        \rowcolor[rgb]{ .949,  .949,  .949}
         &  & {\color{MyRed}\ding{55}} & \ms{77.84}{1.57} & \ms{78.97}{0.22} & \ms{78.01}{1.34} & \ms{77.53}{1.69} & \ms{79.10}{0.28} & \ms{77.84}{1.57} & \ms{77.50}{1.73} \\
        \rowcolor[rgb]{ .949,  .949,  .949}
        \multirow{-7}{*}{V+A+T+C} & \multirow{-2}{*}{\textbf{CVLA (Ours)}} & DY11k & \bms{80.68}{0.49} & \bms{81.11}{0.31} & \bms{80.72}{0.45} & \bms{80.62}{0.53} & \bms{81.16}{0.38} & \bms{80.68}{0.49} & \bms{80.62}{0.54} \\
        \bottomrule
    \end{tabular}
\end{table*}

%% file: table/ur_funny.tex
\begin{table}[!t]
    \small
    \centering
    \caption{\label{tab:ur_funny} Performance comparison on UR-FUNNY dataset. PT denotes Pre-Training.}
    \begin{tabular}{lcccc}
        \toprule
        \multicolumn{1}{c}{\textbf{Approach}} & \textbf{PT} & \textbf{Acc} & \textbf{Mac-F1} & \textbf{Wtd-F1} \\
        \midrule
        C-MFN \cite{conf/emnlp/HasanRZZTMH19} & {\color{MyRed}\ding{55}} & \ms{50.58}{1.48} & \ms{34.00}{0.66} & \ms{34.00}{1.64} \\
        BBFN \cite{conf/icmi/HanCG0MP21} & {\color{MyRed}\ding{55}} & \ms{55.25}{1.85} & \ms{55.00}{1.89} & \ms{55.00}{1.91} \\
        LF-VILA \cite{NEURIPS2022_f8290ccc} & {\color{MyGreen}\ding{51}} & \ms{51.33}{1.35} & \ms{49.23}{1.08} & \ms{49.23}{0.84} \\
        CMHP \cite{conf/ecai/LiuPZSLZ23} & {\color{MyGreen}\ding{51}} & \ms{51.88}{0.64} & \ms{45.19}{4.15} & \ms{45.19}{3.98} \\
        \rowcolor[rgb]{ .949,  .949,  .949}
        \textbf{CVLA (Ours)} & {\color{MyGreen}\ding{51}} & \bms{56.42}{0.69} & \bms{56.22}{0.77} & \bms{56.22}{0.82} \\
        \bottomrule
    \end{tabular}
\end{table}

%% file: table/effect.tex
\begin{table*}[!t]
    \small
    \centering
    \caption{\label{tab:effect_modal} Performance comparison of our CVLA by different modality combinations without and with pre-training.}
    \begin{tabular}{cccccccccc}
        \toprule
        \multirow{2}{*}{\textbf{Modality}} & \multicolumn{3}{c}{\textbf{Without Pre-Training}} & \multicolumn{3}{c}{\textbf{With Pre-Training}} & \multicolumn{3}{c}{\textbf{Improvement}} \\
        \cmidrule(lr){2-4} \cmidrule(lr){5-7} \cmidrule(lr){8-10}
         & \textbf{Acc} & \textbf{Mac-F1} & \textbf{Wtd-F1} & \textbf{Acc} & \textbf{Mac-F1} & \textbf{Wtd-F1} & \textbf{Acc} & \textbf{Mac-F1} & \textbf{Wtd-F1} \\
        \midrule
        V+A & \ms{61.19}{1.62} & \ms{59.56}{2.81} & \ms{59.53}{3.00} & \ms{68.60}{1.31} & \ms{68.26}{1.22} & \ms{68.31}{1.21} & ${\color{MyGreen}\uparrow7.41}$ & ${\color{MyGreen}\uparrow8.70}$ & ${\color{MyGreen}\uparrow8.81}$ \\
        V+T & \ms{67.44}{1.14} & \ms{67.19}{1.04} & \ms{67.12}{1.00} & \ms{68.34}{0.83} & \ms{68.19}{0.89} & \ms{68.24}{0.86} & ${\color{MyGreen}\uparrow0.90}$ & ${\color{MyGreen}\uparrow1.00}$ & ${\color{MyGreen}\uparrow1.12}$ \\
        V+C & \ms{77.04}{0.33} & \ms{76.97}{0.38} & \ms{76.95}{0.40} & \ms{77.36}{0.33} & \ms{77.33}{0.36} & \ms{77.34}{0.36} & ${\color{MyGreen}\uparrow0.32}$ & ${\color{MyGreen}\uparrow0.36}$ & ${\color{MyGreen}\uparrow0.39}$ \\
        A+T & \ms{67.18}{1.49} & \ms{66.54}{1.93} & \ms{66.48}{1.98} & \ms{68.25}{1.20} & \ms{68.04}{1.14}  & \ms{68.03}{1.15} & ${\color{MyGreen}\uparrow1.07}$ & ${\color{MyGreen}\uparrow1.50}$ & ${\color{MyGreen}\uparrow1.55}$ \\
        A+C & \ms{76.97}{0.64} & \ms{76.90}{0.57} & \ms{76.88}{0.55} & \ms{77.07}{0.81} & \ms{77.03}{0.83} & \ms{77.05}{0.82} & ${\color{MyGreen}\uparrow0.10}$ & ${\color{MyGreen}\uparrow0.13}$ & ${\color{MyGreen}\uparrow0.17}$ \\
        T+C & \ms{78.45}{0.79} & \ms{78.34}{0.81} & \ms{78.31}{0.82} & \ms{79.16}{1.94} & \ms{79.08}{1.93} & \ms{79.07}{1.95} & ${\color{MyGreen}\uparrow0.71}$ & ${\color{MyGreen}\uparrow0.74}$ & ${\color{MyGreen}\uparrow0.76}$ \\
        V+A+T & \ms{66.76}{0.63} & \ms{66.28}{0.55} & \ms{66.18}{0.56} & \ms{69.63}{0.48} & \ms{69.61}{0.47} & \ms{69.62}{0.47} & ${\color{MyGreen}\uparrow2.87}$ & ${\color{MyGreen}\uparrow3.33}$ & ${\color{MyGreen}\uparrow3.44}$ \\
        V+A+C & \ms{75.11}{1.76} & \ms{74.96}{1.86} & \ms{74.94}{1.87} & \ms{78.10}{0.67} & \ms{78.01}{0.69} & \ms{78.02}{0.69} & ${\color{MyGreen}\uparrow2.99}$ & ${\color{MyGreen}\uparrow3.05}$ & ${\color{MyGreen}\uparrow3.08}$ \\
        V+T+C & \ms{78.26}{1.91} & \ms{78.10}{2.00} & \ms{78.11}{1.95} & \ms{80.13}{0.58} & \ms{80.07}{0.63} & \ms{80.05}{0.65} & ${\color{MyGreen}\uparrow1.87}$ & ${\color{MyGreen}\uparrow1.97}$ & ${\color{MyGreen}\uparrow1.94}$ \\
        A+T+C & \ms{77.78}{1.40} & \ms{77.62}{1.43} & \ms{77.65}{1.40} & \ms{79.20}{0.51} & \ms{79.08}{0.57} & \ms{79.07}{0.60} & ${\color{MyGreen}\uparrow1.42}$ & ${\color{MyGreen}\uparrow1.46}$ & ${\color{MyGreen}\uparrow1.42}$ \\
        \rowcolor[rgb]{ .949,  .949,  .949}
        V+A+T+C & \ms{77.84}{1.57} & \ms{77.53}{1.69} & \ms{77.50}{1.73} & \bms{80.68}{0.49} & \bms{80.62}{0.53} & \bms{80.62}{0.54} & ${\color{MyGreen}\uparrow2.84}$ & ${\color{MyGreen}\uparrow3.09}$ & ${\color{MyGreen}\uparrow3.12}$ \\
        \bottomrule
    \end{tabular}
\end{table*}

%% file: figure/curve.tex
\begin{figure}[!t]
    \centering
    \includegraphics[width=2.5in,trim=20 0 10 0,clip]{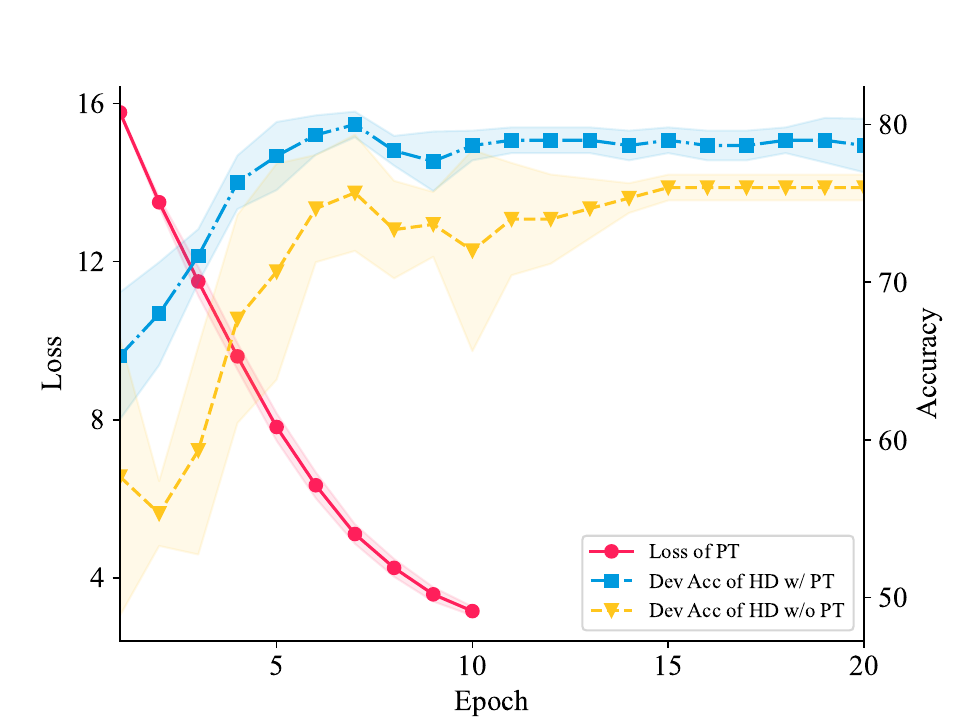}
    \caption{\label{fig:curve} The loss of our pre-training (PT), the accuracy (Acc) of development set (Dev) for humor detection (HD) by fine-tuning after PT and without PT by our CVLA.}
\end{figure}

%% file: table/scale.tex
\begin{table}[!t]
    \small
    \centering
    \caption{\label{tab:effect_scale} Performance comparison with different data scales in pre-training.}
    \begin{tabular}{cccc}
        \toprule
        \textbf{Scale} & \textbf{Acc} & \textbf{Mac-F1} & \textbf{Wtd-F1} \\
        \midrule
        - & \ms{77.84}{1.57} & \ms{77.53}{1.69} & \ms{77.50}{1.73} \\
        3K & \ms{77.30}{0.70} & \ms{77.12}{0.62} & \ms{77.07}{0.61} \\
        5K & \ms{78.58}{1.20} & \ms{78.49}{1.28} & \ms{78.46}{1.30} \\
        8K & \ms{80.42}{1.39} & \ms{80.41}{1.39} & \ms{80.41}{1.39} \\
        \rowcolor[rgb]{ .949,  .949,  .949}
        10K & \bms{80.68}{0.49} & \bms{80.62}{0.53} & \bms{80.62}{0.54} \\
        \bottomrule
    \end{tabular}
\end{table}

%% file: figure/attn.tex
\begin{figure}[!t]
  \centering
  \subfigure[\label{fig:attn_wo_pretrain} Without pre-training.]{
    \includegraphics[width=0.45\linewidth,trim=35 15 30 15,clip]{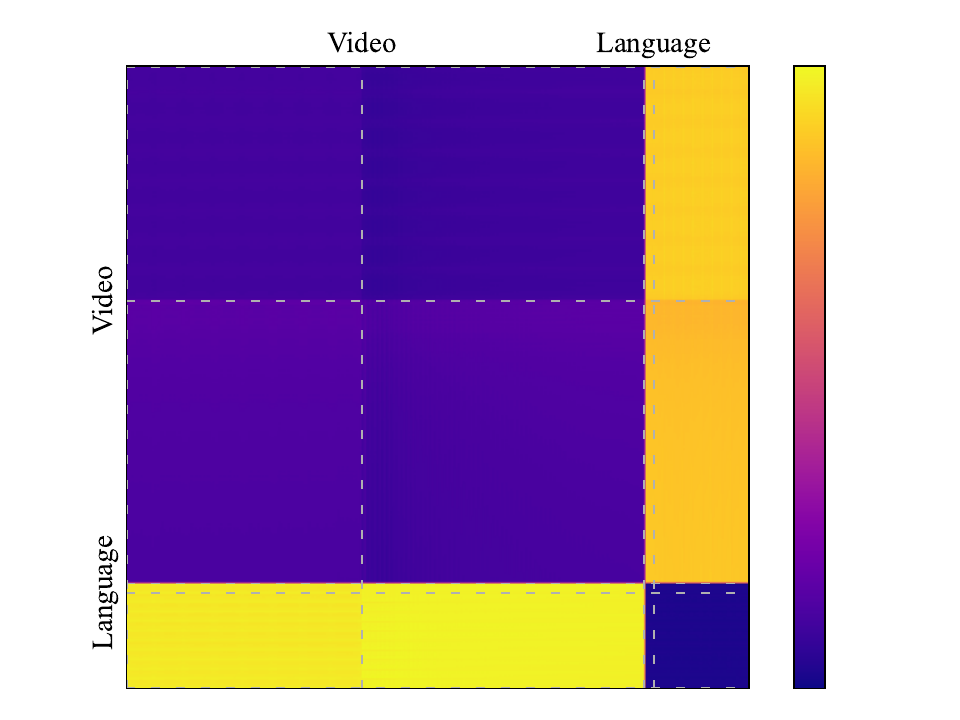}
  }
  \hfill
  \subfigure[\label{fig:attn_finetune} With pre-training.]{
    \includegraphics[width=0.45\linewidth,trim=35 15 30 15,clip]{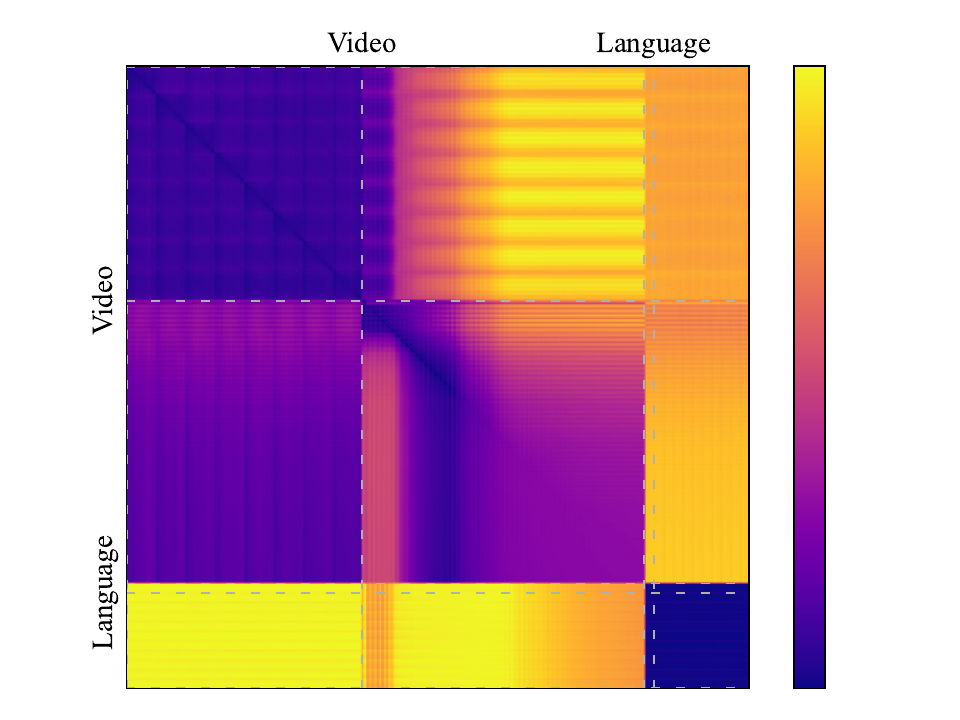}
  }
  \caption{\label{fig:attn} Self-Attention visualization of our multi-modal encoder without and with contrastive pre-training. The four regions separated by gray dashed lines from left to right (top to bottom) represent V, A (video branch), T, C (language branch), respectively.}
\end{figure}

%% file: table/ablation.tex
\begin{table}[!t]
    \small
    \centering
    \caption{\label{tab:effect_contrast} Performance comparison of different ablated approaches from our CVLA.}
    \begin{tabular}{lccc}
        \toprule
        \textbf{Approach} & \textbf{Acc} & \textbf{Mac-F1} & \textbf{Wtd-F1} \\
        \midrule
        \rowcolor[rgb]{ .949,  .949,  .949}
        CVLA & \bms{80.68}{0.49} & \bms{80.62}{0.53} & \bms{80.62}{0.54} \\
        \quad w/o $\textrm{MME}$ & \ms{77.33}{0.32} & \ms{77.21}{0.31} & \ms{77.21}{0.34} \\
        \quad w/o $M^{e}$ & \ms{77.75}{0.45} & \ms{77.57}{0.60} & \ms{77.53}{0.63} \\
        \quad w/o $\mathcal{L}_{V}$ & \ms{79.45}{2.21} & \ms{79.36}{2.28} & \ms{79.33}{2.28} \\
        \quad w/o $\mathcal{L}_{L}$ & \ms{75.72}{1.53} & \ms{75.51}{1.65} & \ms{75.45}{1.69} \\
        \quad w/o $\mathcal{L}_{v^{\prime}}+\mathcal{L}_{l^{\prime}}$ & \ms{79.71}{0.75} & \ms{79.70}{0.76} & \ms{79.69}{0.77} \\
        \bottomrule
    \end{tabular}
\end{table}

%% file: figure/tsne.tex
\begin{figure}[!t]
  \centering
  \subfigure[\label{fig:tsne_wo_pretrain} Without pre-training.]{
    \includegraphics[width=0.45\linewidth,trim=30 20 30 30,clip]{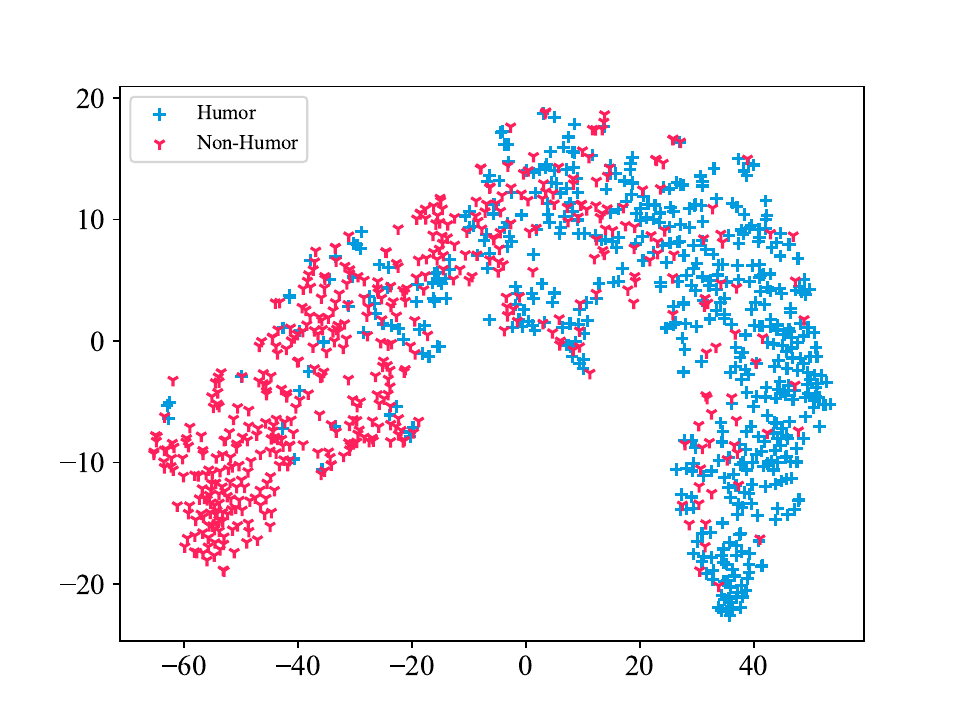}
  }
  \hfill
  \subfigure[\label{fig:tsne_finetune} With pre-training.]{
    \includegraphics[width=0.45\linewidth,trim=30 20 30 30,clip]{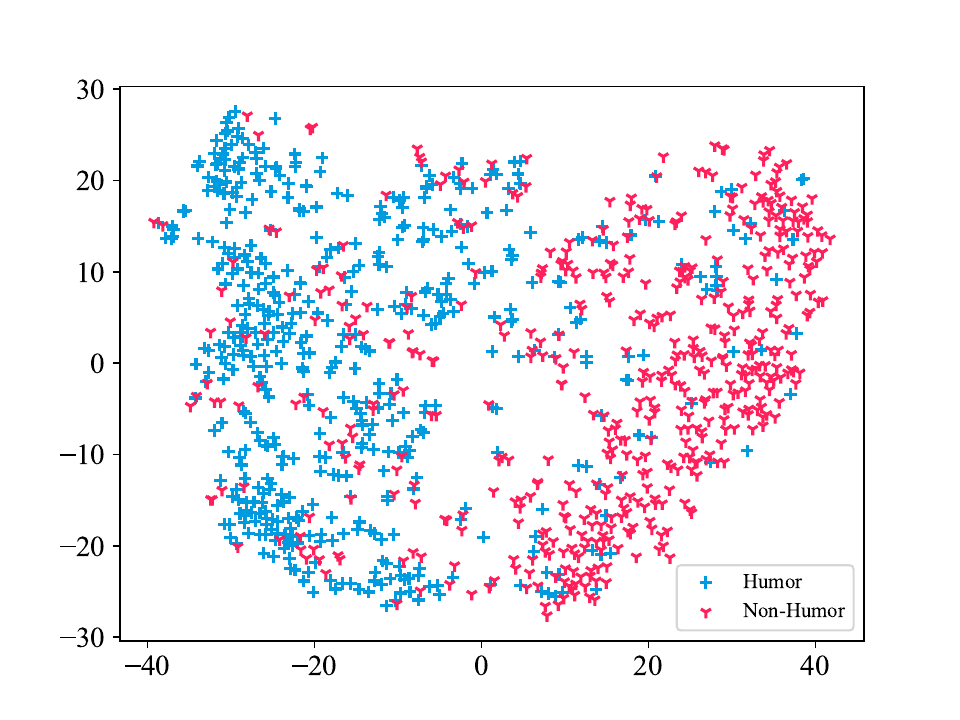}
  }
  \caption{\label{fig:tsne} t-SNE visualization of our multi-modal fusion representations without and with contrastive pre-training.}
\end{figure}

%% file: section/6-conclusion.tex
\section{Conclusion}

In this paper, we concentrate on the short-form video humor detection (SVHD) utilizing commentary data to incorporate the social dimension.
Initially, we develop an expanded multi-modal dataset DY11k from DY24h.
Subsequently, we propose a \textbf{C}omment-aided \textbf{V}ideo-\textbf{L}anguage \textbf{A}lignment (\textbf{CVLA}) approach, employing data-augmented contrastive pre-training to tackle the challenges of SVHD.
We then conduct comparative analyses of CVLA against the state-of-the-art and other competitive baselines on two humor detection datasets, DY11k and UR-FUNNY, to demonstrate the advantages of CVLA.
Furthermore, we provide various interesting analysis to verify the effectiveness of CVLA in details.